\documentclass[letterpaper, 10 pt, conference]{ieeeconf}  
\IEEEoverridecommandlockouts                              
\usepackage{graphicx}
\usepackage{amsmath}
\usepackage{amssymb}
\usepackage{booktabs}
\usepackage{algpseudocode}
\usepackage{algorithm}
\usepackage{multirow}
\newcommand{\argmax}{\operatornamewithlimits{argmax}}
\usepackage{listings}

\usepackage[pagebackref,breaklinks,colorlinks]{hyperref}

\graphicspath{{figures/}}

\begin{document}

\title{PhysicsAssistant: An LLM-Powered Interactive Learning Robot for Physics Lab Investigations}

\author{Ehsan Latif$^1$ \and  Ramviyas Parasuraman$^2$ \and Xiaoming Zhai$^{1,*}$
\thanks{$^1$ AI4STEM Education Center, University of Georgia, Athens, GA 30602, USA}
\thanks{$^2$ School of Computing, University of Georgia, Athens, GA 30602, USA}
\thanks{$^*$ Corresponding author email: \tt{xiaoming.zhai@uga.edu}} 
}
\maketitle

\begin{abstract}
  Robot systems in education can leverage Large language models' (LLMs) natural language understanding capabilities to provide assistance and facilitate learning. This paper proposes a multimodal interactive robot (PhysicsAssistant) built on YOLOv8 object detection, cameras, speech recognition, and chatbot using LLM to provide assistance to students' physics labs. We conduct a user study on ten 8th-grade students to empirically evaluate the performance of PhysicsAssistant with a human expert. The Expert rates the assistants' responses to student queries on a 0-4 scale based on Bloom's taxonomy to provide educational support.
  We have compared the performance of PhysicsAssistant (YOLOv8+GPT-3.5-turbo) with GPT-4 and found that the human expert rating of both systems for factual understanding is same. However, the rating of GPT-4 for conceptual and procedural knowledge (3 and 3.2 vs 2.2 and 2.6, respectively) is significantly higher than PhysicsAssistant (p $<$ 0.05). However, the response time of GPT-4 is significantly higher than PhysicsAssistant (3.54 vs 1.64 sec, p $<$ 0.05). Hence, despite the relatively lower response quality of PhysicsAssistant than GPT-4, it has shown potential for being used as a real-time lab assistant to provide timely responses and can offload teachers' labor to assist with repetitive tasks. To the best of our knowledge, this is the first attempt to build such an interactive multimodal robotic assistant for K-12 science (physics) education.  

\end{abstract}
\section{Introduction}
\label{sec:intro}
In the realm of K-12 education, physics presents a unique set of challenges and opportunities due to the complexity of hands-on labs. Engaging students with the physics theoretical underpinnings and labs is critical, yet often hindered by the abstract nature of the procedures \cite{adeshola2023opportunities}. Students often need individualized and timely assistance to overcome challenges and design solutions \cite{latif2023knowledge}. The use of robots in education offers several advantages: they provide consistent and scalable support, engage students through interactive learning, and can operate in real-time without the need for constant human supervision \cite{chu2022artificial}.

Interactive robots powered by Large Language Models (LLM) like GPT-3.5-turbo can meet this need, bridging the gap between conceptual learning and experiential understanding \cite{vemprala2023chatgpt, ye2023improved, latif2024fine}. LLMs excel in language processing and generating human-like text but struggle with interpreting visual data, a key component in physics education. Models like GPT-3.5-turbo lack multimodal input capabilities, while advanced successors such as GPT-4 \cite{gpt4visioncard, openai2023gptsee, lee2023gemini} support visual, audio, and text inputs but can be expensive and slower for real-time applications \cite{yang2024gpt4tools, lee2023multimodality}. Additionally, concerns about data protection, ethical use, and the need for locally run systems highlight the importance of developing alternative solutions.

To address these limitations, we propose a system that handles multimodal input and provides timely human-like responses efficiently and cost-effectively. Open-sourced deep learning models like YOLOv8 \cite{lou2023dc} offer a viable solution. Despite slightly lower performance compared to GPT-4 for visual understanding, YOLOv8 is cost-effective, efficient, and maintains high performance \cite{zang2024zero}. Its small size (as small as 3.2 million parameters) and ability to accurately identify objects and discern position details make it an excellent candidate for integration with LLMs in educational settings \cite{jin2023target, shen2023ds}.

\begin{figure}[t]
\centering
\begin{center}
 \includegraphics[width=0.97\columnwidth]{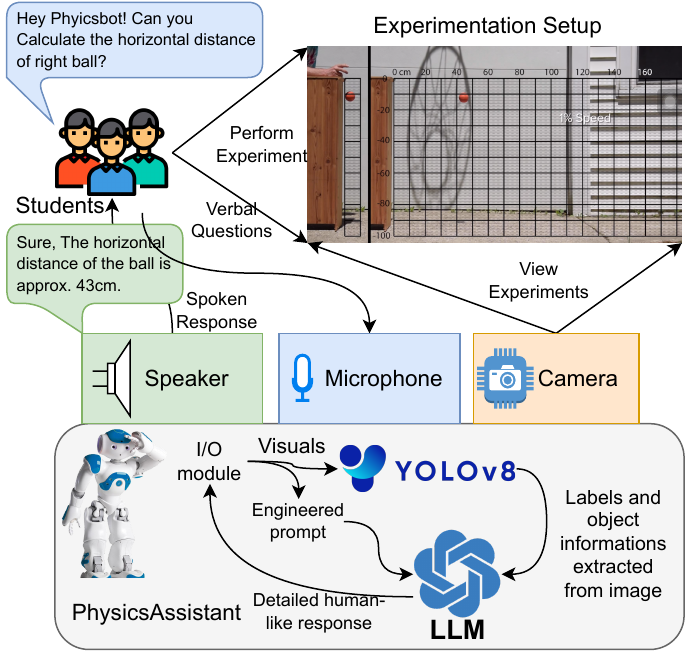}
\end{center}
\vspace{-2mm}
 \caption{An overview of LLM-powered PhysicsAssistant supporting students to perform 8th-grade projectile motion experiment.}
 \label{fig:overview}
 \vspace{-4mm}
\end{figure}

In this paper, we introduce \textit{PhysicsAssistant}, a robot with multimodal processing capabilities (YOLOv8 for visual data, GPT-3.5-turbo for language, and speech processing for audio) designed to assist students in physics lab experiments. The novelty of this system lies in its ability to handle multimodal input using inexpensive and efficient deep-learning models combined with robust LLMs to provide real-time educational support.


The PhysicsAssistant robot is motivated by the urgent need for personalized educational support in K-12 physics labs, where high teacher-to-student ratios hinder individual learning due to students' struggles with complex concepts and the limitations of conventional teaching tools \cite{biswas2023role}. The integration of YOLOv8's object detection with LLMs' conversational intelligence addresses these challenges, revolutionizing physics education by providing an interactive, adaptive, and highly personalized learning experience. This paper demonstrates the potential of the voice-activated PhysicsAssistant robot in a user study, showcasing its ability to engage students and facilitate a deeper and more meaningful learning experience \cite{jangjarat2023public, shafeeg2023voice, latif2024g}. The overview of user-study using PhysicsAssistant can be seen in Fig.~\ref{fig:overview}.

This paper presents the following key contributions:
\begin{itemize}
    \item To the best of our knowledge, this is the first attempt at creating an interactive multimodal robotic assistant empowered with inexpensive and efficient LLM and vision support for real-time K-12 physics educational teaching and learning support. 
    \item We have evaluated the performance of PhysicsAssistant for providing support to students in conducting projectile motion experiments\footnote{\url{https://youtu.be/zyyhCzMWK6Q}}, validated by a human expert on the basis of learning perspectives of Bloom's taxonomy \cite{chandio2016bloom}.
    \item We have compared the performance of PhysicsAssistant (GPT-3.5-turbo for language + YOLOv8 for vision modalities) and GPT-4 (language model with integrated vision modality) and found that PhysicsAssistant provides prompt responses with comparable quality to GPT-4 but is suitable for lab assistant tasks which require timely responses.
    \item We also have observed that human experts have provided above-average ratings for the performance of PhysicsAssistant for real-time applications.
\end{itemize}
Although we specifically focus on the physics education setting, our system can be generalized to science education teaching and learning settings.

\begin{figure}[t]
\centering
\begin{center}
 \includegraphics[width=0.97\columnwidth]{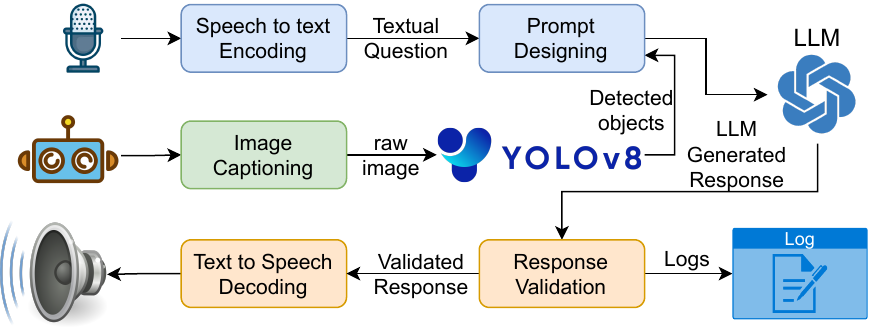}
\end{center}
\vspace{-2mm}
 \caption{PhysicsAssistant's System Architecture, Robot receives audio and visual instructions using an embedded microphone and camera and invokes internal modules for processing, then speaks the response using the inbuilt speaker to students.}
 \label{fig:architecture}
 \vspace{-4mm}
\end{figure}

\section{Related Work}
\label{sec:related_work}

Recent advancements in robotics and Large Language Models (LLMs) have opened new avenues for enhancing Human-Robot Interaction (HRI) and their application in educational settings. Several studies have explored the integration of LLMs like ChatGPT into robotics to facilitate improved HRI \cite{wilcock2023err}, while others have focused on the specific use of robots as educational assistants \cite{donnermann2022investigating}. The common thread among these studies is their emphasis on leveraging technology to enhance learning experiences and outcomes.

Incorporating robots into education has demonstrated significant potential in various studies. Belpaeme et al. \cite{belpaeme2018social} reviewed the impact of social robots in educational settings, highlighting their role in increasing student engagement and motivation. Similarly, Ligthart et al. \cite{ligthart2015selecting} discussed how user attitudes, robot sociability, and embodiment influence user preferences, emphasizing the importance of these factors in educational robots. These studies underscore the importance of HRI in the design and implementation of educational robots.


In the realm of educational robotics, works like those by Cucciniello et al. \cite{cucciniello2022classmate} and Velentza et al. \cite{velentza2021human} have demonstrated the potential of robotic assistants in classrooms. Yet, these approaches often lack the integration of advanced language models and visual perception, limiting their ability to provide contextual and interactive learning support.

The incorporation of visual perception into robotics for education has seen significant advancements with models like YOLOv8 \cite{jin2023target, shen2023ds}. These developments enable robots to better understand and interact with their environment, a crucial aspect of conducting physics experiments. However, existing research primarily focuses on the technical capabilities of these models without integrating them with LLMs for educational purposes.

PhysicsAssistant represents a departure from existing works by merging the conversational intelligence of LLMs with the visual perception capabilities of YOLOv8. This integration allows PhysicsAssistant to not only understand and respond to student inquiries in natural language but also to perceive and interpret the physical setup of experiments \cite{koubaa2023rosgpt, kim2024understanding}. Unlike previous systems, PhysicsAssistant facilitates a real-time, interactive learning experience that closely mirrors human tutoring, addressing the limitations of both solely LLM-powered systems and educational robots without advanced language and visual processing capabilities.

Moreover, our system extends the application of robotics in K-12 education by providing a comprehensive solution that leverages both visual and textual information. This approach not only enhances the robot's ability to assist in complex physics experiments but also represents a novel contribution to the field of educational robotics. Through a user study, we empirically demonstrate PhysicsAssistant's efficacy in supporting physics lab investigations, showcasing its potential to offload teachers' labor and enrich students' learning experiences with higher engagement and understanding.

In addition, studies like those by Ligthart et al. \cite{ligthart2022memory} have shown the importance of personalization and long-term interaction in fostering effective HRI. Memory-based personalization can enhance the learning experience by adapting to the individual needs and preferences of students over time. By integrating these principles, PhysicsAssistant aims to provide a more personalized and effective educational support system.

\section{Proposed PhysicsAssistant System}
\label{sec:system}
The proposed PhysicsAssistant integrates state-of-the-art image recognition, natural language processing, and speech synthesis technologies to provide an interactive educational experience. As depicted in Fig.~\ref{fig:architecture}, the system consists of several interconnected modules working in unison to facilitate real-time physics calculations based on visual input.

The PhysicsAssistant system is designed as a multi-tiered framework that seamlessly processes user interactions from speech input to an informative auditory response. Each tier of the architecture is engineered to handle specific tasks, ensuring efficiency and accuracy. The following subsections detail each component.

\subsection{Speech-to-Text Encoding}

The Speech-to-Text Encoding component uses the Google Speech-To-Text API\footnote{\url{https://cloud.google.com/speech-to-text}} to transcribe student questions in real time. The process can be mathematically represented as:

\begin{equation}
T = f_{\text{decode}}\left(f_{\text{LM}}\left(f_{\text{acoustic}}(S)\right)\right)
\end{equation}

Here, $ S $ denotes the input speech signal. The function $ f_{\text{acoustic}} $ processes this signal to identify phonetic components using a deep neural network (DNN) for noise reduction, normalization, and segmentation into phonemes. The output is then passed through $ f_{\text{LM}} $, which assembles phonemes into words and sentences using a language model. Finally, $ f_{\text{decode}} $ refines the transcription, integrating punctuation based on syntactic analysis and natural pauses, producing the final transcribed text $ T $.

\subsection{Image Processing}

The system captures images using a robotic camera and processes them locally. We use the YOLOv8 model for object detection and image captioning due to its high accuracy and efficiency. Specifically, we employ the yolov8x-pos\footnote{\url{https://docs.ultralytics.com/models/yolov8/}} model, which is an advanced version of the YOLO series, optimized for both speed and precision.

Upon capturing an image, the system processes it through the YOLOv8 model. The detection process involves several key sub-components:
\begin{enumerate}

\item \textbf{Preprocessing}: The image is resized and normalized to match the input requirements of the YOLOv8 model.
\item \textbf{Feature Extraction}: The YOLOv8 model extracts features using convolutional layers, generating a multi-dimensional tensor $\tau$.
\item \textbf{Bounding Box Prediction}: The model predicts bounding boxes $(x, y, w, h)$ for detected objects. This is formalized as:
   \begin{equation}
          B_i = (\sigma(\tau_{i,xy}) + O_{i,xy}) \otimes A_{i,wh}
      \end{equation}

where $\sigma$ is the sigmoid function, $\tau_{i,xy}$ represents the tensor output for the center coordinates, $O_{i,xy}$ are the grid cell offsets, and $A_{i,wh}$ are the anchor box dimensions.
\item \textbf{Classification}: Each bounding box is classified by selecting the class with the highest probability from the softmax-normalized output:
\begin{equation}
   L_i = \argmax (\sigma(\tau_{i,class}))
\end{equation}

where $\tau_{i,class}$ represents the class probability distribution.
 
\end{enumerate}

The parameters for the YOLOv8 model are optimized for high accuracy in object detection and localization. The final output includes bounding boxes and class labels for each detected object.

\begin{figure}[h!]
    \centering
    \includegraphics[width=\linewidth]{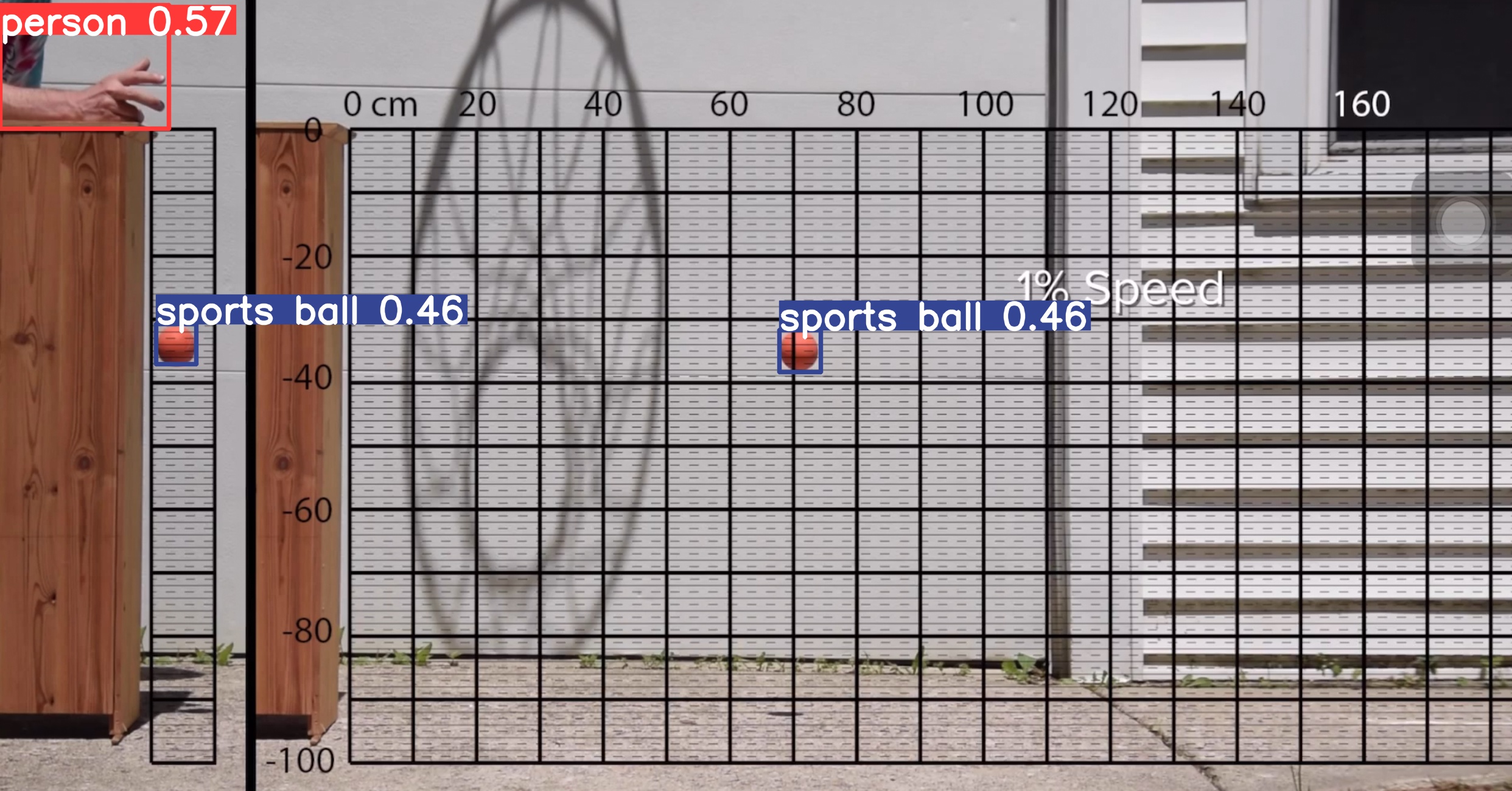}
    \caption{Object detection results using YOLOv8 model}
    \label{fig:result_image}
\end{figure}

The following JSON format provides the caption for the captured image shown in Fig.~\ref{fig:result_image}:

\begin{lstlisting}[basicstyle=\ttfamily\footnotesize, breaklines=true, frame=single]
{
  "image": "capture_image.jpg",
  "objects": [
    {
      "label": "person",
      "confidence": 0.57,
      "bounding_box": [0, 0, 40, 60]
    },
    {
      "label": "sports ball",
      "confidence": 0.46,
      "bounding_box": [30, -20, 50, 10]
    },
    {
      "label": "sports ball",
      "confidence": 0.46,
      "bounding_box": [90, 10, 110, 30]
    }
  ]
}
\end{lstlisting}

\subsection{Prompt Designing}
\label{subsec:prompt_designing}

The Prompt Designing module is a critical intermediary that constructs queries for the LLM based on inputs from the Speech-to-Text Encoding and Image Processing modules. This construction translates visual and verbal inputs into a context-rich prompt that the LLM can interpret accurately.

Once the system has the transcribed text $ T $ from the Speech-to-Text Encoding module and the bounding box and label predictions $ B, L $ from the Image Processing module, a prompt is engineered to reflect the specifics of the physics experiment. The process involves several key steps:

\begin{enumerate}
    \item \textbf{Transforming Detected Objects}: The spatial coordinates and labels of detected objects are translated into a descriptive phrase $ C $:
    \begin{equation}
    C = transform(B, L),
    \end{equation}
    where $ C $ is the image caption created from the bounding box and label predictions $ B, L $.

    \item \textbf{Concatenating Inputs}: The transcribed speech $ T $, the image caption $ C $, and relevant contextual information are combined into a structured prompt. This contextual information includes details about the physics lab experiments, human feedback about the experiment, and a summary of the previous conversation. This can be mathematically described as:
    \begin{equation}
    P = concat(T, \, encode(C),\, Context),
    \end{equation}
    where $ P $ is the final prompt, $ concat $ is the concatenation function, $ encode $ represents the encoding of the image caption, and $ Context $ includes additional pertinent data.

    \item \textbf{System and User Prompts}: The prompt is divided into system and user components to clearly delineate the roles and inputs:
    \begin{itemize}
        \item \textbf{System Prompt}: Provides background information and context about the physics lab setup, previous interactions, and expected response structure.
        \item \textbf{User Prompt}: Includes the transcribed question, the image caption, and additional contextual information.
    \end{itemize}
\end{enumerate}

An example prompt to be fed to GPT-3.5-turbo is as follows:
\begin{lstlisting}[basicstyle=\ttfamily\footnotesize, breaklines=true, frame=single]
System Prompt: 
"Context: This is a physics lab experiment involving projectile motion. The experiment setup includes a person releasing a sports ball. Previous interaction summary: The user asked about the trajectory and distance of the ball in previous questions. Human feedback indicated the need for clear and concise explanations."

User Prompt:
"Question: What is the horizontal distance traveled by the right ball?
Image Caption: The image shows a person with a confidence score of 0.57 and two sports balls with confidence scores of 0.46. The first sports ball is located at coordinates [30, -20, 50, 10] and the second at [90, 10, 110, 30].
Image to World Coordinates: The total world space is 160 x 100, and each grid size is 10x10. The ball can fit in one grid, and its initial position is 0x0.
"
\end{lstlisting}

The Prompt Designing module carefully balances detail and succinctness to ensure the LLM generates accurate and contextually relevant responses, enhancing the interaction quality and overall utility of the PhysicsAssistant system in an educational setting. The system is modular and scalable enough that by changing the context in the system prompt, it can be used for any other lab experiments.

\subsection{LLM Processing}

The generation of responses within the PhysicsAssistant framework utilizes GPT-3.5-turbo, accessible via OpenAI's API\footnote{\url{https://github.com/openai/openai-python}}. This language model (LLM) synthesizes responses $R$ from a given prompt $P$. The underlying process is abstracted as a function $G$, mathematically represented by:

\begin{equation}
R = G(P),
\end{equation}

where $P$ is the input prompt comprising natural language queries, and $R$ is the model-generated response. The function $G$ embodies a convolution of linguistic comprehension, generation mechanisms, and domain-specific knowledge (in this case, physics) to produce responses that are contextually coherent and scientifically substantiated.

\subsubsection{Function $G$ Derivation}
Combining these components, the function $G$ can be derived as the composite operation of tokenization, embedding, sequential processing through Transformer blocks, and response synthesis. This can be encapsulated as:

\begin{equation}
G(P) = \text{Decode}(\text{Transformer}(\text{Embed}(\text{Tokenize}(P)))),
\end{equation}

where the Decode operation synthesizes the response $R$ from the contextualized embeddings produced by the Transformer layers. Through this derivation, it becomes evident that $G$ not only encompasses natural language understanding and generation but also integrates physics knowledge to ensure the generated response $R$ is both contextually relevant and scientifically accurate.

\subsubsection{Perceiving the Prompt and Providing Human-like Feedback}
When the LLM perceives the prompt, it processes the input to understand the context and specific details provided, such as the transcribed question and the encoded image caption. By combining this information with its pre-trained knowledge, the LLM generates a response that mimics human-like feedback, ensuring clarity, coherence, and relevance.

\subsubsection{Example LLM Response}
Given the previously mentioned prompt, the LLM response would be:

\begin{lstlisting}[basicstyle=\ttfamily\footnotesize, breaklines=true, frame=single]
"The horizontal distance traveled by the right ball can be calculated by analyzing the position coordinates and Image to World Coordinates for conversion provided. Given that the right ball's coordinates are [90, 10, 110, 30] starting from [0,0,10,10], we can deduce that the horizontal distance from the release point to its current position is approximately 100 cm."
\end{lstlisting}

This response showcases the LLM's ability to interpret the prompt accurately, utilize contextual information, and provide a clear and precise answer to the user's question, demonstrating its capability to offer human-like feedback in an educational setting.

\subsection{Response Validation}

The Response Validation phase is critical for ensuring the system's outputs are accurate and appropriate. It employs a series of heuristic checks and physics-based verifications, symbolized by a validation function $ V $:

\begin{equation}
V_r = H(R) \land P_v(R),
\end{equation}

where $H_r$ is a boolean reflecting the outcome of linguistic and coherence checks on the response $R$. These checks include grammar, syntax, and semantic consistency evaluations using existing language validation techniques. $P_v(R)$ is a boolean that indicates whether the response $R$ is scientifically accurate and contextually appropriate within the domain of physics. $\land$ denotes the logical AND operator. This composite validation ensures that a response is accepted ($V_r = \text{true}$) only if it successfully passes both heuristic and domain-specific verifications.

\subsubsection{Linguistic and Coherence Checks}
The linguistic validation involves several steps:
\begin{itemize}
    \item \textbf{Grammar Check}: Uses natural language processing (NLP) tools such as spaCy or Grammarly API to ensure the response is grammatically correct.
    \item \textbf{Syntax Check}: Verifies the proper structure of sentences.
    \item \textbf{Semantic Consistency}: Ensures that the response maintains logical coherence and meaning throughout the text. This involves using techniques like Named Entity Recognition (NER) and Dependency Parsing.
\end{itemize}

\subsubsection{Physics-based Verifications}
The domain-specific validation involves:
\begin{itemize}
    \item \textbf{Scientific Accuracy}: Cross-referencing the response with a predefined set of physics rules and formulas relevant to the lab experiment. This can be achieved using rule-based systems or knowledge bases.
    \item \textbf{Contextual Appropriateness}: Ensures that the response is suitable for the specific lab session, considering the experimental setup and previous interactions. This involves contextual analysis using historical data from the lab sessions and feedback from human experts.
\end{itemize}

\subsubsection{Implementation Details}
The validation system integrates these checks using a pipeline approach:
\begin{enumerate}
    \item The response $R$ generated by GPT-3.5-turbo is first passed through the linguistic validation module.
    \item If $H_r$ evaluates to true, the response is then processed by the physics-based validation module.
    \item If both $H_r$ and $P_v(R)$ are true, the response is accepted ($V_r = \text{true}$). If either check fails, $V_r$ is false.
\end{enumerate}

\subsubsection{Response Revision Mechanism}
In the event that $V_r$ evaluates to false, indicating a failure in either heuristic or physics-based checks (or both), the system initiates a response revision mechanism. This mechanism involves:

\begin{itemize}
    \item \textbf{Feedback Incorporation}: Identifying specific deficiencies in the response (e.g., grammatical errors, scientific inaccuracies) and modifying the prompt $P'$ accordingly.
    \item \textbf{Re-invocation of Response Generation}: Re-invoking the response generation function with the revised prompt $P'$ to generate an updated response $R'$:
    \begin{equation}
    R' = G(P'),
    \end{equation}
    where $P'$ incorporates feedback from the validation process.
    \item \textbf{Iterative Refinement}: This iterative refinement cycle continues until a response meets the established criteria for both coherence and scientific accuracy, at which point $V_r$ is set to true.
\end{itemize}



If the response fails any of these checks, the system revises the prompt and regenerates the response, ensuring it meets the necessary standards before presenting it to the user.

This detailed validation process enhances the reliability and educational value of the PhysicsAssistant system, ensuring that responses are not only linguistically correct but also scientifically accurate and contextually relevant.

\subsection{Text-to-Speech Decoding}

In the final stage of the PhysicsAssistant system's workflow, the Text-to-Speech Decoding module is tasked with converting validated textual responses into audible speech using RobuTrans model \cite{li2020robutrans}. The operation of converting textual response $R$ into speech $S_o$ can be mathematically represented as:
\begin{equation}
S_o =  \text{Vocoder}(\text{Decoder}(\text{Encoder}(\mathbf{E}(R)))),
\end{equation}

Where $\mathbf{E}$ is the encoder's operation, $\text{Encoder}$ and $\text{Decoder}$ are layers of self-attention mechanism, $\text{Vocoder}$ is a module to convert mel-spectrograms into waveform audio, and $ S_o $ is the output speech signal.

Through this robust transformer-based model, the PhysicsAssistant system ensures the generation of natural-sounding speech from text, enhancing the user experience by providing auditory feedback that is both clear and contextually accurate.

\subsection{Data Logging}
All processed data, user queries, and system responses are also logged for continuous improvement and audit trails. The architecture is designed with scalability, allowing for future enhancements in AI capabilities and additional modules to handle more complex physics problems and user interactions.



\subsection{Operational Overview}

Figure \ref{fig:overview} provides an operational overview of the PhysicsAssistant system. A user query in the form of speech is transcribed into text and, alongside an image, serves as input for the system. The YOLOv8 object detection framework analyzes the image, detecting objects and their properties, which informs the GPT-3.5-turbo LLM to generate an appropriate response to the query. This response is audibly relayed to the user after passing through a response validation mechanism, ensuring both the pedagogical and factual accuracy of the information provided.

\subsection{Case Study: Ball Trajectory Calculation}

As a practical example, the system can calculate the horizontal distance of a ball's trajectory, as demonstrated in Figure \ref{fig:overview}. Using object detection to determine the ball's position and image captioning to contextualize the scene, the LLM formulates a response based on physical principles. The resultant calculation is then conveyed back to the user through the text-to-speech module, simulating a natural learning interaction.

\section{Experimental Analysis}
\subsection{Setup}
To perform the experiment, we devised a PhysicsAssistant robot with a camera for visual perception, a microphone for acoustic perception, and a speaker to interact with students. The physics experiment focused on projectile motion, where students were tasked with understanding the concept through hands-on activities. The robot was strategically positioned to clearly capture the students' voices and view the experiment. As a voice-activated robot, once invoked by the student by calling, \textit{"Hey PhysicsAssistant!"}, PhysicsAssistant would capture the moment and process the verbal instructions by triggering the call.

To ensure the questions were representative of typical student inquiries during a lab session, we conducted preliminary studies to gather common questions from students. These questions were then refined and validated by educational experts to ensure they covered a broad range of scenarios and learning objectives. This approach aimed to simulate a realistic lab environment where students interact naturally with the PhysicsAssistant.

The visuals were processed through the onboard lightweight YOLOv8 to receive visual details in JSON format, which were then embedded into the engineered prompt used to query the online LLM (GPT-3.5-turbo via API\footnote{\url{https://github.com/openai/openai-python}}). The LLM's responses were converted into voice through text-to-speech. All interactions were logged for longer conversations and further evaluations. We also conducted similar experiments using GPT-4 for comparative analysis using the same prompt format by including encoded images rather than image captions. The experiments were conducted under controlled conditions with IRB\cite{ritchie2021using} trained human experts present to address human-subject-related concerns.

\subsection{Evaluation Criteria}
The evaluation of the PhysicsAssistant system was conducted in an empirical study involving ten students, each interacting with the system by asking a series of five questions at distinct stages of the physics experiment. These stages included the beginning, the first quarter, the midpoint, the third quarter, and the end of the experiment. The questions aimed to evaluate the PhysicsAssistant's ability to make inferences based on the lab scene and address the diverse needs of students:

\begin{enumerate}
    \item What is the horizontal distance traveled by the right ball?
    \item What is the vertical distance traveled by the right ball?
    \item Why do both balls have the same vertical distance but different horizontal distances?
    \item What is the horizontal distance traveled by the right ball when it hits the ground?
    \item If the left ball has less weight, will both balls hit the ground simultaneously?
\end{enumerate}

The provided questions were used to ensure that all key aspects of the experiment were covered, but students were also encouraged to ask spontaneous questions to mimic a real-life lab setting. This combination ensured that the system was tested comprehensively.

\subsubsection{Full System Testing}
To ensure the full system was tested, including speech-to-text and text-to-speech components, we recorded and analyzed the entire interaction flow. This included capturing audio inputs, processing them through the speech-to-text module, and subsequently validating the text responses generated by the LLM before converting them back to speech. Any errors in these processes were noted and factored into the overall evaluation of the system's performance.

\subsection{Human Expert Evaluation}
The responses generated by the PhysicsAssistant were evaluated by a human expert based on a taxonomy adopted from \cite{chandio2016bloom}, which includes four dimensions of knowledge: Factual Knowledge, Conceptual Knowledge, Procedural Knowledge, and Metacognitive Knowledge. 

\textit{Factual Knowledge (FK)} encompasses the basic recall of information, facts, or details. \textit{Conceptual Knowledge (CK)} pertains to the understanding of interrelationships among different facts and concepts. \textit{Procedural Knowledge (PK)} involves the skills and techniques necessary to carry out tasks or solve problems. Lastly, \textit{Metacognitive Knowledge (MK)} covers the awareness and understanding of one's own thought processes and cognitive abilities, which extends beyond content knowledge and encompasses an individual's ability to monitor, control, and reflect on their learning processes.


To evaluate the LLM's performance, a subject specialist human expert was provided with the experimentation logs containing students' questions and the LLM's responses. The human expert rated the LLM's responses for each question on a scale of 0 - 4 (0-Incorrect, 1-Mostly Incorrect, 2-Partially Correct, 3-Mostly Correct, 4-Correct) based on the four dimensions. This comprehensive evaluation ensured that the system's performance was thoroughly assessed, considering all components and potential sources of error.

\section{Results}
The experiment involved evaluating the responses generated by PhysicsAssistant and GPT-4 using a detailed taxonomy for educational objectives, which includes FK, CK, PK, and MK \cite{chandio2016bloom}. The ratings for each question can be seen in Table~\ref{tab:response_ratings}.

\begin{table*}[h!]
\centering
\caption{Comparison of Performance between GPT-3.5-turbo + YOLOv8 and GPT-4}
\label{tab:response_ratings}
\begin{tabular}{|l|cccc|c|cccc|c|}
\hline
\multirow{2}{*}{\textbf{Questions}} & \multicolumn{5}{c|}{\textbf{PhysicsAssistant (GPT-3.5-turbo + YOLOv8)}} & \multicolumn{5}{c|}{\textbf{GPT-4}}                                \\ \cline{2-11} 
& Factual & Conceptual & Procedural & Meta- & \multirow{2}{*}{Average} & Factual & Conceptual & Procedural & Meta- & \multirow{2}{*}{Average} \\
& Knowledge & Knowledge  & Knowledge  & cognitive &                          & Knowledge & Knowledge  & Knowledge  & cognitive & \\ \hline
Q1 & 4 & 3 & 3 & 3 & 3.25 & 4 & 4 & 4 & 4 & 4 \\
Q2 & 4 & 3  & 4 & 4 & 3.75 & 4 & 3 & 4 & 3 & 3.5 \\
Q3 & 4 & 2 & 3 & 3 & 3 & 4 & 3 & 3 & 4 & 3.5 \\
Q4 & 4 & 2 & 2 & 3 & 2.75 & 4 & 3 & 3 & 3 & 3.25 \\
Q5 & 3 & 1 & 1 & 2 & 1.75 & 3 & 2 & 2 & 2 & 2.25 \\ \hline
Average & 3.8 & 2.2 & 2.6 & 3 & 2.9 & 3.8  & 3 & 3.2 & 3.4 & 3.2 \\ \hline
\end{tabular}
\vspace{-0.5cm}
\end{table*}
\subsection{Expert Rating}

The overall performance of the PhysicsAssistant was commendable, with an average rating of 2.9 out of 4 across all dimensions and questions. The system demonstrated strong Factual Knowledge (FK) with a rating of 3.8, driven by the high object detection accuracy (86\%) of YOLOv8. A paired samples t-test comparing PhysicsAssistant (GPT-3.5-turbo + YOLOv8) with GPT-4 in FK showed no statistically significant difference, $ t(4, 49) = -1.00, p = .374 $. This indicates that while we cannot conclude the performances are comparable, more data is needed to determine if there is a difference.




The lowest average rating was observed in Q5 (1.75), which required advanced conceptual and procedural reasoning. This suggests a need for strategies like chain-of-thought prompting to improve higher-order thinking skills.

A question-by-question t-test comparison revealed a significant difference for Q1, $ t(4, 49) = -3.00, p = .058 $, indicating the need for improvements in addressing higher-order thinking skills. For Q2 to Q5, no significant differences were found ($ p = .391, .182, .638, $ and $ .182 $, respectively), showing consistent performance across these questions.

While PhysicsAssistant shows promise in providing educational assistance, particularly in tasks requiring factual recall, further development is needed to address deeper conceptual and procedural questions. This includes prompt engineering and strategies like chain-of-thought prompting to enhance the system's reasoning skills.

It was observed that the LLM-based system, with prompt engineering, could reliably identify FK, even in brief student responses. CK was discerned when the LLM connected elements of FK within its responses. However, evaluating MK posed challenges, as the LLM tended to hallucinate without explicit cues in the reference material.

\subsection{Efficiency}
The efficiency of PhysicsAssistant was also evaluated in terms of response time, which is crucial for maintaining the flow of interaction in an educational setting. The response time for PhysicsAssistant and GPT-4 was measured from the initiation of the student's query to the delivery of the system's verbal response. This includes the time taken for speech-to-text conversion, processing, and text-to-speech output, which was approximately 100ms for both the compared LLMs and is included in the overall response times.

PhysicsAssistant's response time additionally comprised the duration taken by YOLOv8 for image segmentation detection and labeling, the conversion of raw data to a descriptive prompt, and the query and response time of GPT-3.5-turbo. In contrast, GPT-4's response time included the time taken for the image to bit-string conversion, embedding to prompt, and subsequent query and response generation. 

A paired samples t-test was conducted to compare the response time between (excluding initial and final processing time (~100ms), which is the same for both) PhysicsAssistant (GPT-3.5-turbo + YOLOv8) and GPT-4 across all questions. The results revealed a statistically significant difference in the response time, $ t(4) = -6.847, p = .002 < 0.05 $, with PhysicsAssistant demonstrating significantly faster responses than GPT-4 (see Table~\ref{table:response_time}).

The following table provides the response times for both systems across all questions:

\begin{table}[h!]
\centering
\caption{Response Time Comparison between PhysicsAssistant and GPT-4}
\label{table:response_time}
\begin{tabular}{|c|c|c|c|c|}
\hline
\multirow{2}{*}{\textbf{Questions}} & \multicolumn{3}{|c|}{\textbf{PhysicsAssistant (sec)}} & \multirow{2}{*}{\textbf{GPT-4 (sec)}}\\
\cline{2-4} 
 & YOLOv8 & GPT-3.5-turbo & Total & \\ \hline
Q1 & 0.3 & 0.8 & \textbf{1.1} & 2.1 \\
Q2 & 0.2 & 1.5 & \textbf{1.7} & 3.4 \\
Q3 & 0.3 & 1.2 & \textbf{1.5} &3.8 \\
Q4 & 0.4 & 1.4 & \textbf{1.8} & 4.1 \\
Q5 & 0.4 & 1.7 & \textbf{2.1} & 4.3 \\ \hline
Average & 0.32 & 1.32 & \textbf{1.64}  & 3.54 \\ \hline
\end{tabular}
\end{table}

The average response time of PhysicsAssistant was 1.64 $\pm$ 0.24 (0.32 sec of YOLOv8 + 1.32 sec of GPT-3.5-turbo) seconds, while GPT-4's average response time was 3.54 $\pm$ 0.87 seconds The significant reduction in response time by PhysicsAssistant not only improves the user experience by providing quicker feedback and suggests the potential for a more dynamic and fluid interaction in educational environments, which is essential for maintaining student engagement during physics lab sessions. This improved efficiency can be attributed to the local processing capabilities of YOLOv8 for image segmentation and the avoidance of the time-consuming process of image-to-bit-string conversion and data transfer over the network required by GPT-4.

\section{Discussion}

This study presented the development and evaluation of PhysicsAssistant, an interactive robotic system designed to support K-12 students in physics laboratories. The integration of GPT-3.5-turbo with YOLOv8 for object detection and scene interpretation has demonstrated a novel approach in educational robotics, particularly within the context of STEM education.

The potential of PhysicsAssistant lies in its ability to provide real-time assistance by combining the strengths of LLMs in natural language processing with the precision of object detection algorithms. Our results indicate that PhysicsAssistant excels in factual knowledge retrieval and has considerable promise in aiding procedural and metacognitive tasks.


PhysicsAssistant's contributions to the field of human-robot interaction are significant. Demonstrating the practical application of an LLM-powered robot in a physics lab sets a precedent for future research in HRI, particularly in educational settings. Its success in a physics lab setting heralds a future where robots not only assist in teaching but also enhance learning by making it more accessible, interactive, and personalized for K-12 students.


\section{Conclusion}
The PhysicsAssistant system exhibits a promising capability to support K -12 education in physics by integrating state-of-the-art object detection and advanced LLMs. The empirical results highlight its competence in providing assistance with accurate factual and procedural responses. Despite this, challenges remain in addressing complex conceptual questions and those pertaining to higher-order metacognitive skills. The findings suggest that while the system effectively aids students with immediate, information-based queries, developing deeper conceptual understanding and advanced reasoning skills may benefit from enhanced LLM prompting strategies and domain-specific fine-tuning. Future work will aim to refine these aspects and foster a more robust educational tool that answers questions and promotes comprehensive learning and critical thinking.

\bibliographystyle{IEEEtran}
\bibliography{refs}

\begin{thebibliography}{10}
\providecommand{\url}[1]{#1}
\csname url@samestyle\endcsname
\providecommand{\newblock}{\relax}
\providecommand{\bibinfo}[2]{#2}
\providecommand{\BIBentrySTDinterwordspacing}{\spaceskip=0pt\relax}
\providecommand{\BIBentryALTinterwordstretchfactor}{4}
\providecommand{\BIBentryALTinterwordspacing}{\spaceskip=\fontdimen2\font plus
\BIBentryALTinterwordstretchfactor\fontdimen3\font minus \fontdimen4\font\relax}
\providecommand{\BIBforeignlanguage}[2]{{%
\expandafter\ifx\csname l@#1\endcsname\relax
\typeout{** WARNING: IEEEtran.bst: No hyphenation pattern has been}%
\typeout{** loaded for the language `#1'. Using the pattern for}%
\typeout{** the default language instead.}%
\else
\language=\csname l@#1\endcsname
\fi
#2}}
\providecommand{\BIBdecl}{\relax}
\BIBdecl

\bibitem{adeshola2023opportunities}
I.~Adeshola and A.~P. Adepoju, ``The opportunities and challenges of chatgpt in education,'' \emph{Interactive Learning Environments}, pp. 1--14, 2023.

\bibitem{latif2023knowledge}
E.~Latif, L.~Fang, P.~Ma, and X.~Zhai, ``Knowledge distillation of llm for education,'' \emph{arXiv preprint arXiv:2312.15842}, 2023.

\bibitem{chu2022artificial}
S.-T. Chu, G.-J. Hwang, and Y.-F. Tu, ``Artificial intelligence-based robots in education: A systematic review of selected ssci publications,'' \emph{Computers and education: Artificial intelligence}, vol.~3, p. 100091, 2022.

\bibitem{vemprala2023chatgpt}
S.~Vemprala, R.~Bonatti, A.~Bucker, and A.~Kapoor, ``Chatgpt for robotics: Design principles and model abilities,'' \emph{Microsoft Auton. Syst. Robot. Res}, vol.~2, p.~20, 2023.

\bibitem{ye2023improved}
Y.~Ye, H.~You, and J.~Du, ``Improved trust in human-robot collaboration with chatgpt,'' \emph{IEEE Access}, 2023.

\bibitem{latif2024fine}
E.~Latif and X.~Zhai, ``Fine-tuning chatgpt for automatic scoring,'' \emph{Computers and Education: Artificial Intelligence}, p. 100210, 2024.

\bibitem{gpt4visioncard}
{Open AI}, ``Gpt-4v(ision) system card.'' September 2023, published on September 25, 2023.

\bibitem{openai2023gptsee}
------, ``Chatgpt can now see, hear, and speak,'' \url{https://openai.com/blog/chatgpt-can-now-see-hear-and-speak}, September 2023, published on September 25, 2023.

\bibitem{lee2023gemini}
G.-G. Lee, E.~Latif, L.~Shi, and X.~Zhai, ``Gemini pro defeated by gpt-4v: Evidence from education,'' \emph{arXiv preprint arXiv:2401.08660}, 2023.

\bibitem{yang2024gpt4tools}
R.~Yang, L.~Song, Y.~Li, S.~Zhao, Y.~Ge, X.~Li, and Y.~Shan, ``Gpt4tools: Teaching large language model to use tools via self-instruction,'' \emph{Advances in Neural Information Processing Systems}, vol.~36, 2024.

\bibitem{lee2023multimodality}
G.-G. Lee, L.~Shi, E.~Latif, Y.~Gao, A.~Bewersdorf, M.~Nyaaba, S.~Guo, Z.~Wu, Z.~Liu, H.~Wang \emph{et~al.}, ``Multimodality of ai for education: Towards artificial general intelligence,'' \emph{arXiv preprint arXiv:2312.06037}, 2023.

\bibitem{lou2023dc}
H.~Lou, X.~Duan, J.~Guo, H.~Liu, J.~Gu, L.~Bi, and H.~Chen, ``Dc-yolov8: Small-size object detection algorithm based on camera sensor,'' \emph{Electronics}, vol.~12, no.~10, p. 2323, 2023.

\bibitem{zang2024zero}
Z.~Zang, C.~Lin, C.~Tang, T.~Wang, and J.~Lv, ``Zero-shot aerial object detection with visual description regularization,'' \emph{arXiv preprint arXiv:2402.18233}, 2024.

\bibitem{jin2023target}
Y.~Jin, Z.~Shi, X.~Xu, G.~Wu, H.~Li, and S.~Wen, ``Target localization and grasping of nao robot based on yolov8 network and monocular ranging,'' \emph{Electronics}, vol.~12, no.~18, p. 3981, 2023.

\bibitem{shen2023ds}
L.~Shen, B.~Lang, and Z.~Song, ``Ds-yolov8-based object detection method for remote sensing images,'' \emph{IEEE Access}, vol.~11, pp. 125\,122--125\,137, 2023.

\bibitem{biswas2023role}
S.~Biswas, ``Role of chat gpt in education,'' \emph{Available at SSRN 4369981}, 2023.

\bibitem{jangjarat2023public}
K.~Jangjarat, T.~Kraiwanit, P.~Limna, and R.~Sonsuphap, ``Public perceptions towards chatgpt as the robo-assistant,'' \emph{Jangjarat, K., Kraiwanit, T., Limna, P., \& Sonsuphap}, 2023.

\bibitem{shafeeg2023voice}
A.~Shafeeg, I.~Shazhaev, D.~Mihaylov, A.~Tularov, and I.~Shazhaev, ``Voice assistant integrated with chat gpt,'' \emph{Indonesian Journal of Computer Science}, vol.~12, no.~1, 2023.

\bibitem{latif2024g}
E.~Latif, G.-G. Lee, K.~Neuman, T.~Kastorff, and X.~Zhai, ``G-sciedbert: A contextualized llm for science assessment tasks in german,'' \emph{arXiv preprint arXiv:2402.06584}, 2024.

\bibitem{chandio2016bloom}
M.~T. Chandio, S.~M. Pandhiani, and R.~Iqbal, ``Bloom's taxonomy: Improving assessment and teaching-learning process.'' \emph{Journal of education and educational development}, vol.~3, no.~2, pp. 203--221, 2016.

\bibitem{wilcock2023err}
G.~Wilcock and K.~Jokinen, ``To err is robotic; to earn trust, divine: Comparing chatgpt and knowledge graphs for hri,'' in \emph{2023 32nd IEEE International Conference on Robot and Human Interactive Communication (RO-MAN)}.\hskip 1em plus 0.5em minus 0.4em\relax IEEE, 2023, pp. 1396--1401.

\bibitem{donnermann2022investigating}
M.~Donnermann, P.~Schaper, and B.~Lugrin, ``Investigating adaptive robot tutoring in a long-term interaction in higher education,'' in \emph{2022 31st IEEE International Conference on Robot and Human Interactive Communication (RO-MAN)}.\hskip 1em plus 0.5em minus 0.4em\relax IEEE, 2022, pp. 171--178.

\bibitem{belpaeme2018social}
T.~Belpaeme, J.~Kennedy, A.~Ramachandran, B.~Scassellati, and F.~Tanaka, ``Social robots for education: A review. science robotics 3, 21 (2018),'' 2018.

\bibitem{ligthart2015selecting}
M.~Ligthart and K.~P. Truong, ``Selecting the right robot: Influence of user attitude, robot sociability and embodiment on user preferences,'' in \emph{2015 24th IEEE International Symposium on Robot and Human Interactive Communication (RO-MAN)}.\hskip 1em plus 0.5em minus 0.4em\relax IEEE, 2015, pp. 682--687.

\bibitem{cucciniello2022classmate}
I.~Cucciniello, G.~L’Arco, A.~Rossi, C.~Autorino, G.~Santoro, and S.~Rossi, ``Classmate robot: A robot to support teaching and learning activities in schools,'' in \emph{2022 31st IEEE International Conference on Robot and Human Interactive Communication (RO-MAN)}.\hskip 1em plus 0.5em minus 0.4em\relax IEEE, 2022, pp. 906--911.

\bibitem{velentza2021human}
A.-M. Velentza, N.~Fachantidis, and I.~Lefkos, ``Human or robot university tutor? future teachers’ attitudes and learning outcomes,'' in \emph{2021 30th IEEE International Conference on Robot \& Human Interactive Communication (RO-MAN)}.\hskip 1em plus 0.5em minus 0.4em\relax IEEE, 2021, pp. 236--242.

\bibitem{koubaa2023rosgpt}
A.~Koubaa, ``Rosgpt: Next-generation human-robot interaction with chatgpt and ros,'' 2023.

\bibitem{kim2024understanding}
C.~Y. Kim, C.~P. Lee, and B.~Mutlu, ``Understanding large-language model (llm)-powered human-robot interaction,'' \emph{arXiv preprint arXiv:2401.03217}, 2024.

\bibitem{ligthart2022memory}
M.~E. Ligthart, M.~A. Neerincx, and K.~V. Hindriks, ``Memory-based personalization for fostering a long-term child-robot relationship,'' in \emph{2022 17th ACM/IEEE International Conference on Human-Robot Interaction (HRI)}.\hskip 1em plus 0.5em minus 0.4em\relax IEEE, 2022, pp. 80--89.

\bibitem{li2020robutrans}
N.~Li, Y.~Liu, Y.~Wu, S.~Liu, S.~Zhao, and M.~Liu, ``Robutrans: A robust transformer-based text-to-speech model,'' in \emph{Proceedings of the AAAI conference on artificial intelligence}, vol.~34, no.~05, 2020, pp. 8228--8235.

\bibitem{ritchie2021using}
K.~L. Ritchie, ``Using irb protocols to teach ethical principles for research and everyday life: A high-impact practice.'' \emph{Journal of the Scholarship of Teaching and Learning}, vol.~21, no.~1, pp. 120--130, 2021.

\end{thebibliography}

\end{document}